\documentclass[10pt,journal,compsoc]{IEEEtran}
\ifCLASSOPTIONcompsoc
  \usepackage[nocompress]{cite}
\else
  \usepackage{cite}
\fi
\usepackage{amsmath,amssymb,graphicx}
\usepackage{booktabs}
\usepackage{threeparttable}
\usepackage{multirow}
\usepackage{subfigure}
\usepackage{array}
\graphicspath{{figure/}}

\newcommand{\tabincell}[2]{
    \begin{tabular}{@{}#1@{}}#2\end{tabular}
}

\ifCLASSINFOpdf
\else
\fi
\begin{document}
%
\title{Impacts of Solar Irradiance and Wind Speed on District Heat System}
%
%
%
%

\author{Jiyang Xie, Zhanyu Ma, and Jun Guo%
\IEEEcompsocitemizethanks{\IEEEcompsocthanksitem J. Xie, Z. Ma and J. Guo are with Pattern Recognition and Intelligent Systems Lab., Beijing University of Posts and Telecommunications, China.\protect\\

}
}

%
%

\markboth{ }%
{Xie \MakeLowercase{\textit{et al.}}: Impacts of Whether Conditions on District Heat System}
%



\IEEEtitleabstractindextext{%
\begin{abstract}
Using artificial neural network for the prediction of heat demand has attracted more and more attention. Weather conditions, such as ambient temperature, wind speed and direct solar irradiance, have been identified as key input parameters. In order to further improve the model accuracy, it is of great importance to understand the influence of different parameters. Based on an Elman neural network (ENN), this paper investigates the impact of direct solar irradiance and wind speed on predicting the heat demand of a district heating network. Results show that including wind speed can generally result in a lower overall mean absolute percentage error (MAPE) (6.43\%) than including direct solar irradiance (6.47\%); while including direct solar irradiance can achieve a lower maximum absolute deviation (71.8\%) than including wind speed (81.53\%). In addition, even though including both wind speed and direct solar irradiance shows the best overall performance (MAPE=6.35\%), ENN may not always benefit from the simultaneous introduction of both wind speed and direct solar irradiance, according to MAPE in broken down ranges of heat demands (0$\sim$150MW, 150$\sim$300MW, 300$\sim$450MW and $>$450MW).
\end{abstract}

}

\maketitle

\IEEEdisplaynontitleabstractindextext

%
\IEEEpeerreviewmaketitle

\IEEEraisesectionheading{\section{Introduction}\label{sec:introduction}}

%
%
%
%
\IEEEPARstart{E}{ven} though district heating (DH) has been considered as the most efficient, environment friendly and cost-effective method for supplying heat to buildings, it is nowadays facing big challenges of further improving efficiency, reducing cost and enhancing profitability. In general, reducing the peak load has been considered as a crucial issue regarding both efficiency improvement and cost saving. In order to reduce the peak, it is of great importance to understand the consumption pattern and predict the peak load precisely. Meanwhile, examining the experience in the electricity market shows that demand response is an effective way to save energy and reduce the cost \cite{li15,pan16,wang14}. For the time-critical DH system \cite{Basantaval16}, similar benefits can also be achieved if heat can be supplied in a more dynamic way, such as hourly, according to the demand. To adjust the heat supply, accurate prediction of heat demand is essential for DH companies.

Thermal energy is the product of mass flow, temperature difference, the specific heat of the water, and time. Models for predicting the heat demand can be classified into two categories: physical models, which are developed based on energy balance according to the principle of heat transfer \cite{melikyan07,fouda10}; and statistic models, which are built up based on measured data and data processing technologies \cite{torchio09,dotzauer02,ma14}. Evolving technologies about smart meters and smart energy network open up new opportunities for the second category as more and more energy data become available \cite{sun15}. Data analysis can extract valuable information from large amount of data, which can be further used for model design and algorithm implementation \cite{ma17,shao16}. Many data analysis methods \cite{ma17} have been developed and adopted in the intelligent energy networks. Therefore, the statistic models attract more and more attention in the prediction of heat demand, due to its unique advantages, such as the ability to reflect the sociological behaviors of consumers. Generally, the most popular methods for data analysis in energy networks are linear regression (LR), support vector machine (SVM) and neural networks. Correspondingly, many statistic models have been developed to predict heat demand and shown good abilities to produce accurate predictions \cite{ma17,shamshirband15}.

Statistic models need to correlate the heat demand to some parameters, such as meteorological parameters, the building type and time of the day \cite{ma14} etc. The ambient temperature is the most important meteorological parameter, since it determines the temperature difference between indoor and outdoor, which is the main driving force for heat transfer \cite{dotzauer02,shamshirband15}. However, it is not sufficient to obtain satisfying results by only considering ambient temperature. In order to further improve the accuracy of prediction, other meteorological parameters, such as wind speed and direct solar irradiance, have also been considered. For example, Michalakakou et al. \cite{michalakakou02} involved direct solar irradiance as one of the inputs in an artificial neural network (ANN) model to forecast the heat demand in residential buildings. Yang et al. \cite{yang15} combined the numerical weather prediction (NWP) with an ANN model for the projection of heat load, in which both direct solar irradiance and wind speed were taken as inputs. Kusiak et al. \cite{kusiak10} identified wind speed as an important parameter in predicting building energy demand. Fu et al. \cite{fu16} verified the importance of direct solar irradiance on the thermal load of a micro DH network. It can be concluded that both the wind speed and the direct solar irradiance have been identified as important meteorological parameters, which should be considered as inputs to a statistic model. However, the impacts of direct solar irradiance and wind speed have not been quantitatively compared regarding the prediction of energy consumption. Different opinions exist in the open literature about which parameter should give a higher priority \cite{edwards12,jain14,chramcov11}. In order to achieve a high accuracy of prediction, it is of great significance to clarify their quantitative impacts. Therefore, this work aims to quantitatively investigate the impacts of direct solar irradiance and wind speed on heat demand prediction.

Meanwhile, most of the previous studies mainly focused on the heat demand from the consumption side, for example the heat demand of different types of buildings; while fewer efforts have been dedicated from the production side, due to lack of data \cite{idowu14,gadd13,nielsen06,kato08}. However, an accurate prediction of heat demand is more significant to district heating companies as the dynamic heat demand is needed by the planning and optimization of heat production and the determination of heat price. Therefore, the contributions of this work to academia and industry include: (I) to determine the quantitative influence of direct solar irradiation and wind speed; (II) to identify the most important parameter regarding the improvement of the prediction accuracy; and (III) to develop a statistic model with high accuracy for the prediction of heat demand at DH plants.

\begin{table}[htbp]
    \caption{Nomenclature}
    \label{tab:nomenclature}
    \vspace{-4mm}
    \centering
    \begin{tabular}{ll}
        \toprule
        Symbol&Quantity\\
        \midrule
        DH&district heating\\
        ENN&Elman neural network\\
        MAPE&mean absolute percentage error\\
        DMAPE&daily mean absolute percentage error\\
        RMSE&root mean square error\\
        MAD&maximum absolute deviation\\
        MW&mega watt\\
        $u(t)$&input of input layer\\
        $x(t)$&output of hidden layer\\
        $x_c(t)$&output of context layer\\
        $y(t)$&output of output layer\\
        $y_d(t)$&training target\\
        $W$&connection weight matrix between hidden layers\\
        $W^c$&\tabincell{l}{connection weight matrix between hidden layer and \\the context layer}\\
        $f(\cdot)$&tangent sigmoid transfer function\\
        $g(\cdot)$&linear transfer function\\
        $\eta$&learning rate of $W$\\
        $\eta_c$&learning rate of $W^c$\\
        \bottomrule
    \end{tabular}
\end{table}

\section{Model Description}\label{sec:model}

Statistic Models are developed based on Elman neural networks (ENN) in this paper.  ENN is an effective method for predicting energy demands. In previous studies, it has been used in short term heat demand prediction \cite{fu16}, mid-long term electricity demand prediction \cite{zhang13} and host energy demand prediction \cite{huang12}.

\subsection{Elman Neural Network}\label{ssec:enn}

ENN is a global feed forward and local recurrent neural network proposed in 1990 by Elman for the purpose of overcoming the drawbacks of the traditional neural networks \cite{elman90}. An ENN generally comprises four levels: the input, the hidden, the context, and the output layers. The structures of input, hidden, and output layers are similar to the normal feedforward neural network. The role of context layer nodes is to store the output of the hidden layer nodes, which are used for activating the nodes of hidden layer in the next time step and equivalent to the time delay operator.

\begin{figure}[htbp]
    \centering
    \includegraphics[width=0.95\linewidth]{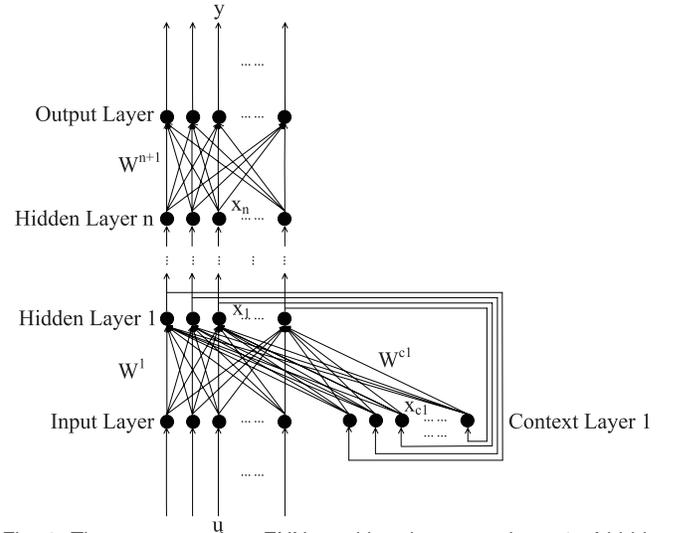}
    \vspace{-5mm}
    \caption{The structure of an ENN-m with only context layer 1 of hidden layer 1 showed as an example.}
    \label{fig:ennmstructure}
\end{figure}

A typical ENN has one hidden layer with delayed feedback. However, an ENN with multiple hidden layers (ENN-m) is often used in order to achieve a better prediction performance, which is represented as:
\begin{equation}
    x_1(t)=f[W^1u(t)+W^{c,1} x_{c,1}(t)],\label{eq:enninput}
\end{equation}
\begin{equation}
    x_i(t)=f[W^ix_{i-1}(t)+W^{c,i}x_{c,i}(t)],i=2,3,\cdots,n,\label{eq:ennhidden}
\end{equation}
\begin{equation}
    x_{c,i}(t)=x_i(t-1),i=1,2,\cdots,n,\label{eq:enncontext}
\end{equation}
\begin{equation}
    y(t)=g[W^{n+1}x_n(t)],\label{eq:ennoutput}
\end{equation}
\noindent where $t$ is the time step, $n$ is the number of hidden layers, $u(t)$ is the input of the model, $x_{c,i}(t)$ and $x_i(t)$ are the output of context layer $i$ and hidden layer $i$, $y(t)$ is the output of output layer, $W^1$ is the connection weight matrix between the input layer and the hidden layer $1$, $W^i$ is the connection weight matrix between the hidden layer $i$ and the hidden layer $(i-1)$, $W^{n+1}$ is the connection weight matrix between the hidden layer $n$ and the output layer, and $W^{c,i}$ is the connection weight matrix between the hidden layer $i$ and the context layer $i$ respectively. $f(\cdot)$ and $g(\cdot)$ are transfer functions, of which $f(\cdot)$ is a tangent sigmoid transfer function and $g(\cdot)$ is a linear transfer function. The structure of ENN-m is shown in Figure \ref{fig:ennmstructure}. It is noted that Figure \ref{fig:ennmstructure} describes the structure of ENN-m with n hidden layers but only shows context layer $1$ for hidden layer $1$ as an example.

\subsection{Data}\label{ssec:data}

Hourly data during the period of 2008-2011 were collected from a utility company in V\"aster\"as, Sweden, including the heat consumption, ambient temperature, direct solar irradiance and wind speed. This DH company provides about 1.6TWh heat to more than 14 thousand users, which can be categorized into six groups: villas, multi-family buildings, office \& school buildings; commercial buildings; hospital/social services buildings; and industry buildings. Considering the clear difference about the heat demand in working days and non-working days, this work mainly focuses on working days in order to achieve a higher accuracy.

To quantitatively evaluate the impacts of wind speed and direct solar irradiance on the prediction of heat demand, respectively, four datasets were created for ENN model training, which details are listed in Table \ref{tab:datasetdescription}.

\begin{table}[htbp]
    \caption{Dataset Description}
    \label{tab:datasetdescription}
    \vspace{-4mm}
    \centering
    \begin{tabular}{ccccc}
        \toprule
        Dataset&\tabincell{c}{Heat \\demand}&\tabincell{c}{Ambient \\temperature}&\tabincell{c}{Direct solar \\irradiance}&\tabincell{c}{Wind \\speed}\\
        \midrule
        Dataset A&$\surd$&$\surd$&$\times$&$\times$\\
        Dataset B&$\surd$&$\surd$&$\surd$&$\times$\\
        Dataset C&$\surd$&$\surd$&$\times$&$\surd$\\
        Dataset D&$\surd$&$\surd$&$\surd$&$\surd$\\
        \bottomrule
    \end{tabular}
\end{table}

\subsection{Model Training}\label{ssec:training}

\subsubsection{Training Steps of Models}\label{sssec:trainingstep}

For ENN-m, the length of slide window and the number of hidden layers of neural networks are two key parameters. In order to achieve a better performance, their impacts on the prediction of heat demand are evaluated. To investigate the impact of the length of slide windows, data in consecutive 2, 4 and 8 hours are combined to create a super-vector. A super-vector is the input of neural network at one time step. A step size, which is set as a half of the length of slide window, is selected to update the super-vector. For example, if 4 hours are chosen as the length of slide window, then the step size will be 2 hours. Correspondingly, the first super-vector will contain the data from the 1st to the 4th hour and the second super-vector contains the data from the 3rd to the 6th hour. To investigate the impact of the number of hidden layers of ENN, 4 and 8 layers are tested.

The training steps of ENN are as follows:
\begin{eqnarray}
    &\left\{
    \begin{array}{l}
    \Delta W_{ij}^{n+1}=\eta_{n+1}\delta_i^{n+1}x_{nj}(t)\\
    \Delta W_{jl}^p=\eta_p\delta_j^px_{pj}(t)\\
    \Delta W_{jq}^1=\eta_1\delta_j^1u_q(t-1)\\
    \Delta W_{jl}^{c,p}=\eta_{c,p}\delta_j^px_{pl}(t-1)
    \end{array}
    \right.,\\
    &i=1,2,\cdots,m; j=1,2,\cdots,s; l=1,2,\cdots,s;\nonumber\\
    &p=2,3,\cdots,n; q=1,2,\cdots,r,\nonumber\label{eq:trainingweights}
\end{eqnarray}
\begin{equation}
    \delta_i^{n+1}=(y_{d,i}(t)-y_i(t))g'(\cdot),\label{eq:derouttohid}
\end{equation}
\begin{equation}
    \delta_j^p=\Sigma_{i=1}^m(\delta_i^{p+1}W_{ij}^{p+1})f'(\cdot), p=1,2,\cdots,n,\label{eq:derhidtohid}
\end{equation}

\noindent where $\eta_p$  and $\eta_{cp}$ ($p=1,2,\cdots,n$) are the learning rates of  $W^p$ and $W^{cp}$, $m$ is the number of the output layer nodes, $s$ is the number of the hidden layer nodes, $r$ is the number of the input layer nodes, $n$ is the number of the hidden layers, $y_{d,i}(t)$ means the target corresponding $y_i(t)$, and $f'(\cdot)$ and $g'(\cdot)$ are the derivatives of $f(\cdot)$ and $g(\cdot)$ in section \ref{ssec:enn}.

15 hidden nodes and 1 output node are chosen for each hidden layer and the output layer, respectively. The number of input nodes is equal to (the length of slide window + the number of factors). For example, if the length of slide window is 4 hours and dataset D, which includes 3 factors, is chosen, the number of input nodes will be 7 (4+3).

The Z-Score or named standard score is used as the normalization method before model training to preprocess measured data, which is represented as follows:
\begin{equation}
    \hat{x}_i=\frac{x_i-\mu}{\sqrt{\sigma^2}}, i=1,2,\cdots,N,\label{eq:normalization}
\end{equation}
\begin{equation}
    \mu=\frac{1}{N}\sum_{i=1}^Nx_i,\label{eq:mean}
\end{equation}
\begin{equation}
    \sigma^2=\frac{1}{N-1}\sum_{i=1}^N(x_i-\mu)^2,\label{eq:variance}
\end{equation}
\noindent where $x_i$ is an element of the measured data,  $x=(x_1,\cdots,x_N)$, $\hat{x}_i$ is an element of the normalized data $\hat{x}=(\hat{x}_1,\cdots,\hat{x}_N)$, $\mu$ and $\sigma^2$ are the sample mean and the unbiased sample variance of $x$, and N is the length of $x$ and $\hat{x}$.

\subsubsection{Model Validation}\label{sssec:validation}

To evaluate the model performance, the mean absolute percentage error (MAPE) and the root mean square error (RMSE) are used as indicators. MAPE is defined as
\begin{equation}
    \text{MAPE}=\frac{1}{N}\sum_{i=1}^N\frac{\left|y_i-y_{pi}\right|}{y_i}\times100\%.\label{eq:mape}
\end{equation}
\noindent RMSE is an absolute indicator to measure the performance of models
\begin{equation}
    \text{RMSE}=\sqrt{\frac{1}{N}\sum_{i=1}^N(y_i-y_{pi})^2}\label{eq:rmse}
\end{equation}
\noindent where $y_i$ is the actual heat demand, $y_{pi}$ is the predicted result, and $N$ is the number of the points of heat demand predictions.

The Student's \emph{t}-test is a statistical hypothesis test and usually used to determine if MAPEs of the models are significantly different from each other. A significant difference on MAPEs means that the factors may play a decisive role regarding the model performance. Meanwhile, a boxplot graphically depicts the distribution of MAPEs. The interquartile range and mid-range of the boxplot are used to find a group of results which have the most concentrated distribution and the lowest quartiles of MAPEs.

\section{Results}\label{sec:results}

\subsection{The impacts of the number of hidden layers and the length of slide window}\label{ssec:impactofhidandwin}

Short-term predictions of heat demand were implemented to investigate the impacts of the number of hidden layers and the length of slide window. The data from October 2008 to February 2009 (five months) were used for model training and those in March 2009 (one month) were used for model validation. To study the impacts of the number of hidden layers and the length of slide window, 6 models were obtained, as shown in Table \ref{tab:modelname}. Here, ``window'' means the length of slide window and ``layers'' denotes the number of hidden layers. Each model was trained by using dataset A-D, respectively. Correspondingly, the inputs for each model are ambient temperature (T$_{ambient}$), T$_{ambient}$ + direct solar irradiance data (R$_{solar irradiance}$), T$_{ambient}$ + wind speed (V$_{wind}$), and T$_{ambient}$ + R$_{solar irradiance}$ + V$_{wind}$. Each model was trained and tested for 10 times in order to analyze the distribution of errors.

\begin{table*}[htbp]
    \caption{Different Models for the Short-term Prediction of Heat Demand}
    \label{tab:modelname}
    \vspace{-4mm}
    \centering
    \begin{tabular}{cccc}
        \toprule
        \tabincell{c}{Length of slide\\windows (hr)}&2&4&8\\
        \midrule
        4 hidden layers&\tabincell{c}{``Elman, window=2, \\layers=4''}&\tabincell{c}{``Elman, window=4, \\layers=4''}&\tabincell{c}{``Elman, window=8, \\layers=4''}\\
        8 hidden layers&\tabincell{c}{``Elman, window=2, \\layers=8''}&\tabincell{c}{``Elman, window=4, \\layers=8''}&\tabincell{c}{``Elman, window=8, \\layers=8''}\\
        \bottomrule
    \end{tabular}
\end{table*}

240 groups of results were obtained from each model. Table \ref{tab:meanvarofmape} and Table \ref{tab:meanvarofrmse} show the mean and the variance of MAPE and RMSE. The computing methods of mean and variance are showed in \eqref{eq:mean} and \eqref{eq:variance}. The smallest mean and variance of MAPE and RMSE of the obtained models have been highlighted.

\begin{table*}[htbp]
    \centering
    \caption{Mean and Variance of MAPE}
    \label{tab:meanvarofmape}
    \vspace{-4mm}
    \begin{tabular}{cc|cc|cc|cc}
        \hline
        \multicolumn{2}{c|}{Length of slide windows (hr)}&\multicolumn{2}{c|}{2}&\multicolumn{2}{c|}{4}&\multicolumn{2}{c}{8}\\
        \hline
        \multicolumn{2}{c|}{Number of hidden layers}&4&8&4&8&4&8\\
        \hline
        \multirow{2}{2cm}{Dataset A}&Mean&\textbf{4.11\%}&4.17\%&4.14\%&4.14\%&4.33\%&4.28\%\\
        &Variance&1.39E-06&1.70E-06&2.20E-07&\textbf{1.01E-07}&7.40E-07&6.61E-07\\
        \hline
        \multirow{2}{2cm}{Dataset B}&Mean&\textbf{4.21\%}&4.26\%&4.34\%&4.33\%&4.52\%&4.50\%\\
        &Variance&2.30E-06&1.98E-06&4.50E-07&\textbf{3.48E-07}&8.33E-07&1.96E-06\\
        \hline
        \multirow{2}{2cm}{Dataset C}&Mean&\textbf{4.17\%}&4.23\%&4.18\%&4.20\%&4.46\%&4.44\%\\
        &Variance&2.40E-06&1.91E-06&\textbf{8.99E-08}&2.30E-07&9.39E-07&6.21E-07\\
        \hline
        \multirow{2}{2cm}{Dataset D}&Mean&4.37\%&4.35\%&\textbf{4.32}\%&4.32\%&4.66\%&4.60\%\\
        &Variance&1.89E-06&3.19E-06&4.51E-07&\textbf{1.35E-07}&1.40E-06&1.12E-06\\
        \hline
    \end{tabular}
\end{table*}

\begin{table*}[htbp]
    \centering
    \caption{Mean and Variance of RMSE}
    \label{tab:meanvarofrmse}
    \vspace{-4mm}
    \begin{tabular}{cc|cc|cc|cc}
        \hline
        \multicolumn{2}{c|}{Length of slide windows (hr)}&\multicolumn{2}{c|}{2}&\multicolumn{2}{c|}{4}&\multicolumn{2}{c}{8}\\
        \hline
        \multicolumn{2}{c|}{Number of hidden layers}&4&8&4&8&4&8\\
        \hline
        \multirow{2}{2cm}{Dataset A}&Mean&\textbf{17.8545}&18.2153&18.1363&18.1395&18.2800&18.1169\\
        &Variance&0.0794&0.1134&0.0703&0.0683&0.1059&\textbf{0.0458}\\
        \hline
        \multirow{2}{2cm}{Dataset B}&Mean&\textbf{18.4278}&18.7067&18.9075&18.8704&19.2753&19.2306\\
        &Variance&0.2878&0.1941&0.1265&\textbf{0.0569}&0.1279&0.3684\\
        \hline
        \multirow{2}{2cm}{Dataset C}&Mean&18.3824&\textbf{18.2272}&18.3412&18.3981&18.7891&18.8458\\
        &Variance&0.1337&0.1957&\textbf{0.0447}&0.0539&0.1486&0.0452\\
        \hline
        \multirow{2}{2cm}{Dataset D}&Mean&18.9048&\textbf{18.7553}&18.8673&18.8945&19.6771&19.4505\\
        &Variance&0.1318&0.2483&0.0473&\textbf{0.0272}&0.1770&0.1735\\
        \hline
    \end{tabular}
\end{table*}

Table \ref{tab:diffwindowp} and \ref{tab:diffhidlayerp} show the \emph{p}-value of the Student's \emph{t}-test, where the statistical significance threshold is 0.05 ($\alpha=0.05$). During the test, either the length of slide windows or the number of hidden layers is varied, while the other remains the same. For example, to do the Student's \emph{t}-test between the MAPE of models which have different numbers of hidden layers, the length of slide windows should be kept the same. If the calculated \emph{p}-value is below the chosen threshold, the null hypothesis is rejected in favor to the alternative hypothesis.

\begin{table*}[htbp]
    \centering
    \caption{\emph{P}-value of the Student's \emph{T}-test between MAPE where Lengths of Slide Windows Are 2, 4 and 8 Hrs}
    \label{tab:diffwindowp}
    \vspace{-4mm}
    \begin{tabular}{cccccc}
        \hline
        \multirow{2}{*}{\tabincell{c}{Number of\\hidden layers}}&\multirow{2}{*}{\tabincell{c}{Length of slide\\windows (hr)}}&\multicolumn{4}{c}{Dataset used}\\
        &&Dataset A&Dataset B&Dataset C&Dataset D\\
        \hline
        \multirow{3}{*}{4}&2 and 4&0.426484&0.026466&0.860613&0.242179\\
        &2 and 8&0.000134&2.89E-05&0.000101&8.61E-05\\
        &4 and 8&8.46E-06&7.03E-05&7.75E-08&2.12E-07\\
        \hline
        \multirow{3}{*}{8}&2 and 4&0.498705&0.159898&0.626957&0.596684\\
        &2 and 8&0.033244&0.001101&0.000534&0.001374\\
        &4 and 8&7.13E-05&0.002073&2.29E-07&3.10E-07\\
        \hline
    \end{tabular}
\end{table*}

\begin{table}[htbp]
    \centering
    \caption{\emph{P}-value of the Student's \emph{T}-test between MAPE where Numbers of Hidden Layers Are 4 and 8}
    \label{tab:diffhidlayerp}
    \vspace{-4mm}
    \begin{tabular}{ccccc}
        \hline
        \multirow{2}{*}{\tabincell{c}{Length of slide\\windows}}&\multicolumn{4}{c}{Dataset used}\\
        &Dataset A&Dataset B&Dataset C&Dataset D\\
        \hline
        2&0.339876&0.523546&0.389471&0.784251\\
        4&0.682536&0.63984&0.160953&0.753935\\
        8&0.149755&0.649153&0.668691&0.24389\\
        \hline
    \end{tabular}
\end{table}

%

Comparing the models which have different numbers of hidden layers, those with 8 hidden layers only have slightly lower MAPE and RMSE than those with 4 hidden layers. meanwhile, all the results of the Student¡¯s \emph{t}-test show statistically insignificantly different, which implies that increasing the number of hidden layers is not an effective way to improve the performance. However, the computation time is much longer when the number of hidden layers increases. For example, for a computer with Intel core i5-6600@3.30GHz CPU and 4GB RAM, training a neural network with 16 or more hidden layers consumers more than 5 hours than training a neural network with 8 hidden layers. Therefore, 8 hidden layers are selected in this work.

Comparing the models which have different lengths of slide window, those with a 4-hour slide window have smaller MAPE and RMSE than those with a 8-hour slide window. Results of the Student¡¯s \emph{t}-test also show that the distributions of MAPEs obtained by using 2-hour and 8-hour slide windows have significantly statistical difference. It is the same for the distributions of MAPEs obtained by using 4-hour and 8-hour hour slide windows. However, the distributions of MAPEs obtained by 2 and 4-hour slide window don¡¯t have statistically significant difference. Therefore, the 4-hour slide window is chosen.

\subsection{The Impacts of Data Amount on the Prediction of Heat Demand}\label{ssec:impactofdataamount}

Long-term predictions of heat demand were implemented to investigate the impacts of data amount one year (2010), two year (2009 and 2010) and three year data (2008 to 2010) were used for model training respectively, resulting in three models, ENN-1, ENN-2 and ENN-3.  To evaluate the model performance, the data of 2011 were used for model validation.

Table \ref{tab:meanvarofmapelong} and Table \ref{tab:meanvarofrmselong} show the mean and variance of MAPE and RMSE. The smallest ones have been highlighted in bold. In general, using more data for model training can clearly improve the model performance, no matter what the input factors are.

\begin{table}[htbp]
    \centering
    \caption{Means and Variances of MAPE in Long-term Heat Demand Prediction}
    \label{tab:meanvarofmapelong}
    \vspace{-4mm}
    \begin{tabular}{ccccc}
        \hline
        \multicolumn{2}{c}{Model}&ENN-1&ENN-2&ENN-3\\
        \hline
        \multirow{2}{*}{Dataset A}&Mean&6.63\%&6.45\%&\textbf{6.40\%}\\
        &Variance&6.51E-07&\textbf{7.21E-08}&1.91E-07\\
        \multirow{2}{*}{Dataset B}&Mean&6.61\%&6.51\%&\textbf{6.42\%}\\
        &Variance&2.78E-07&6.82E-07&\textbf{8.64E-08}\\
        \multirow{2}{*}{Dataset C}&Mean&6.67\%&6.48\%&\textbf{6.46\%}\\
        &Variance&\textbf{2.17E-07}&2.28E-07&1.59E-06\\
        \multirow{2}{*}{Dataset D}&Mean&6.68\%&6.51\%&\textbf{6.43\%}\\
        &Variance&2.94E-07&3.44E-07&\textbf{1.99E-07}\\
        \hline
    \end{tabular}
\end{table}

\begin{table}[htbp]
    \centering
    \caption{Means and Variances of RMSE in Long-term Heat Demand Prediction}
    \label{tab:meanvarofrmselong}
    \vspace{-4mm}
    \begin{tabular}{ccccc}
        \hline
        \multicolumn{2}{c}{Model}&ENN-1&ENN-2&ENN-3\\
        \hline
        \multirow{2}{*}{Dataset A}&Mean&15.1301&\textbf{14.7204}&14.7526\\
        &Variance&0.0077&0.0032&\textbf{0.0012}\\
        \multirow{2}{*}{Dataset B}&Mean&15.0065&14.7069&\textbf{14.6842}\\
        &Variance&\textbf{0.0019}&0.0022&0.0024\\
        \multirow{2}{*}{Dataset C}&Mean&14.9463&\textbf{14.5583}&14.6030\\
        &Variance&0.0075&\textbf{0.0022}&0.0067\\
        \multirow{2}{*}{Dataset D}&Mean&14.8834&\textbf{14.5491}&14.5609\\
        &Variance&0.0111&0.0062&\textbf{0.0038}\\
        \hline
    \end{tabular}
\end{table}

\begin{figure}[htb]
    \begin{minipage}[b]{0.48\linewidth}
      \centering
      \centerline{\includegraphics[width=0.95\linewidth]{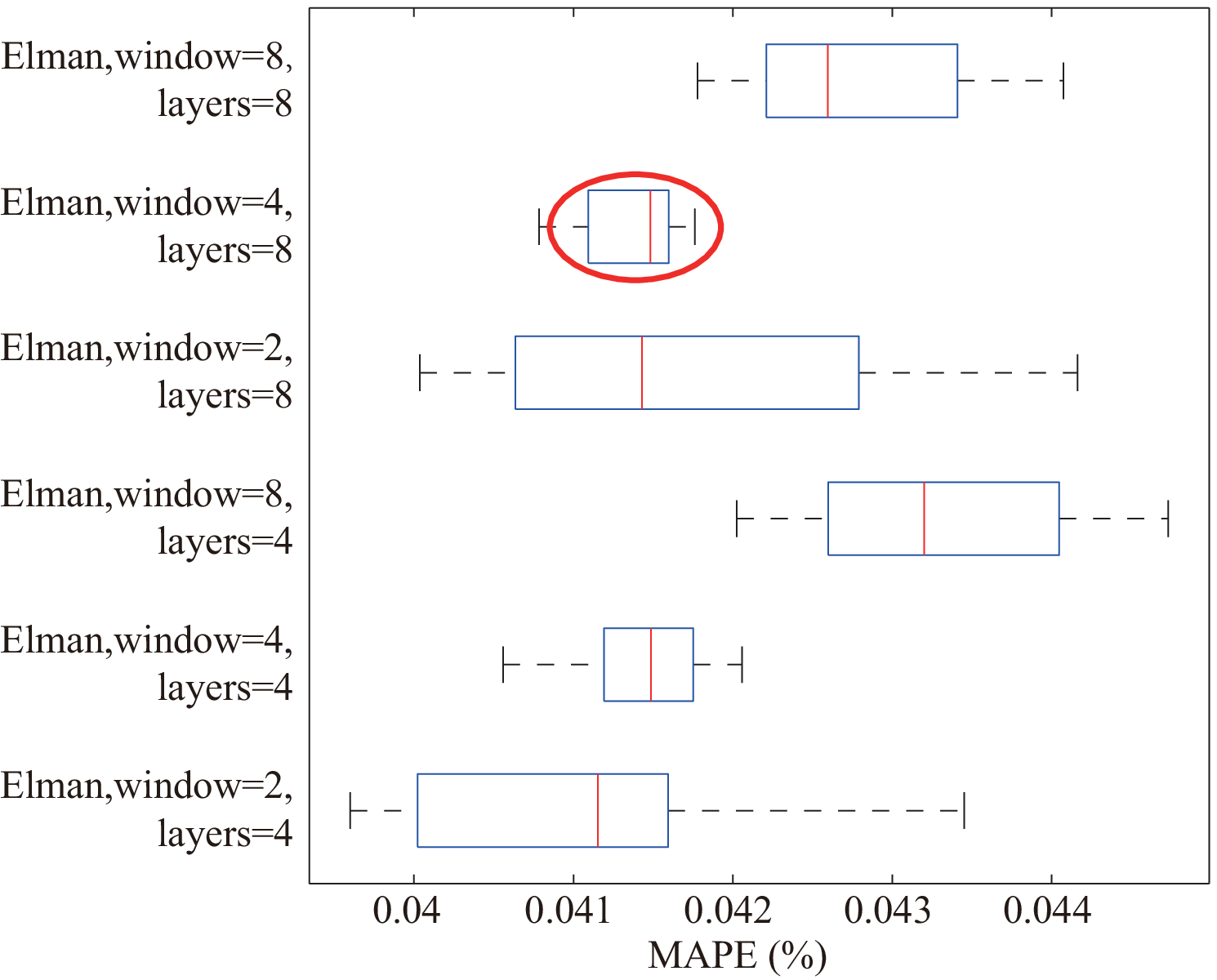}}
      \centerline{\tabincell{c}{(a) Models trained \\by using Dataset A}}\medskip
    \end{minipage}
    \hfill
    \begin{minipage}[b]{0.48\linewidth}
      \centering
      \centerline{\includegraphics[width=0.95\linewidth]{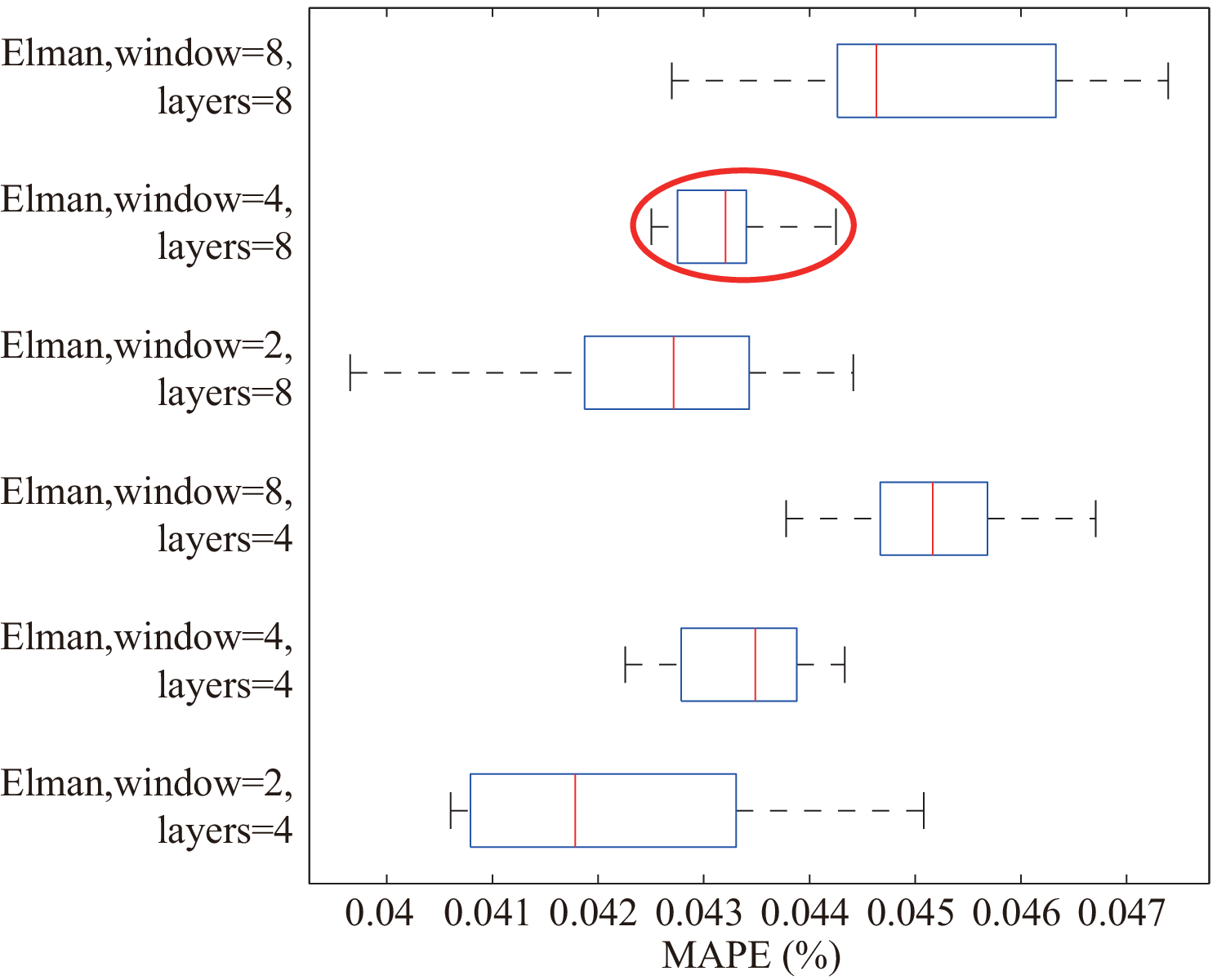}}
      \centerline{\tabincell{c}{(b) Models trained \\by using Dataset B}}\medskip
    \end{minipage}
    \begin{minipage}[b]{0.48\linewidth}
      \centering
      \centerline{\includegraphics[width=0.95\linewidth]{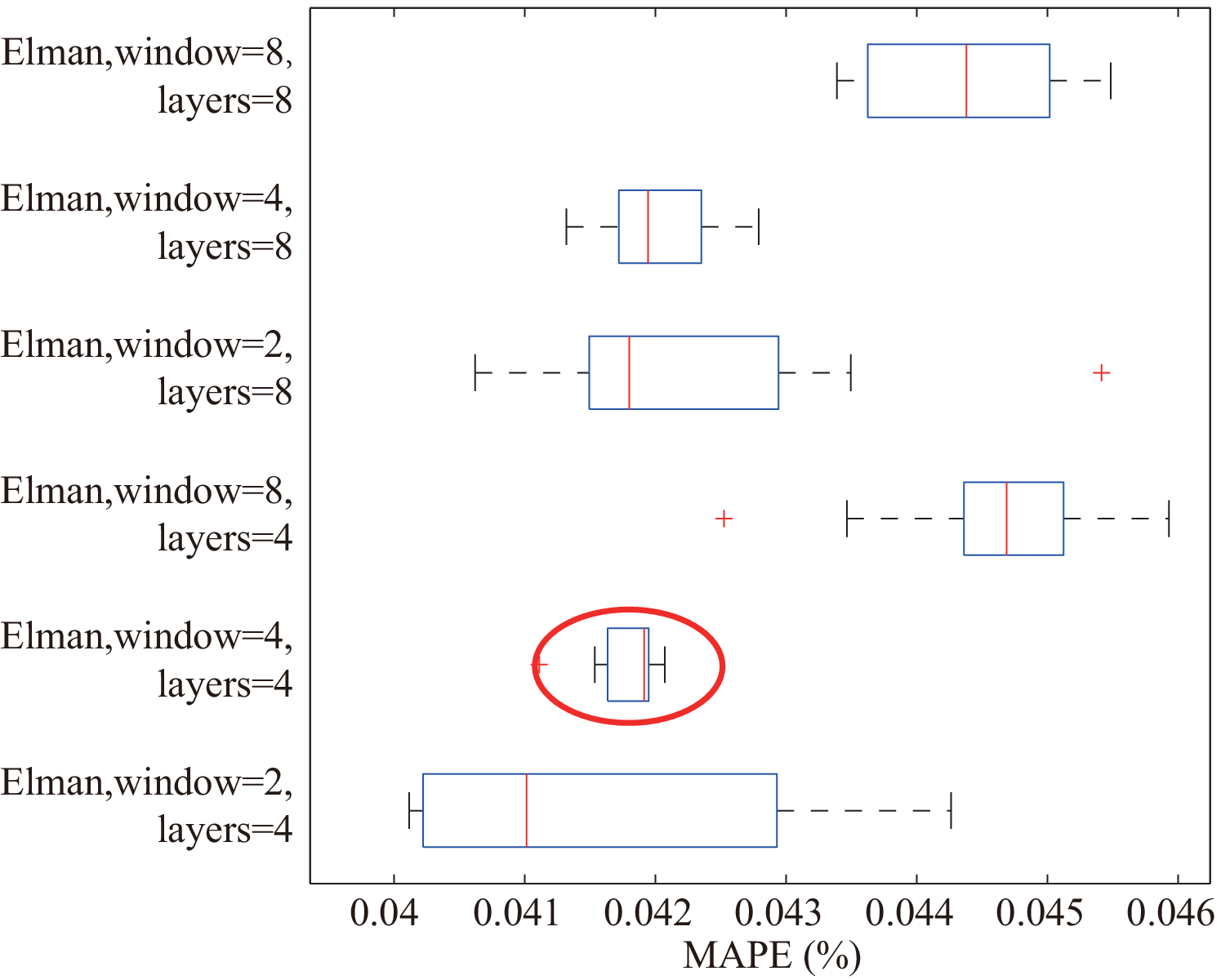}}
      \centerline{\tabincell{c}{(c) Models trained \\by using Dataset C}}\medskip
    \end{minipage}
    \hfill
    \begin{minipage}[b]{0.48\linewidth}
      \centering
      \centerline{\includegraphics[width=0.95\linewidth]{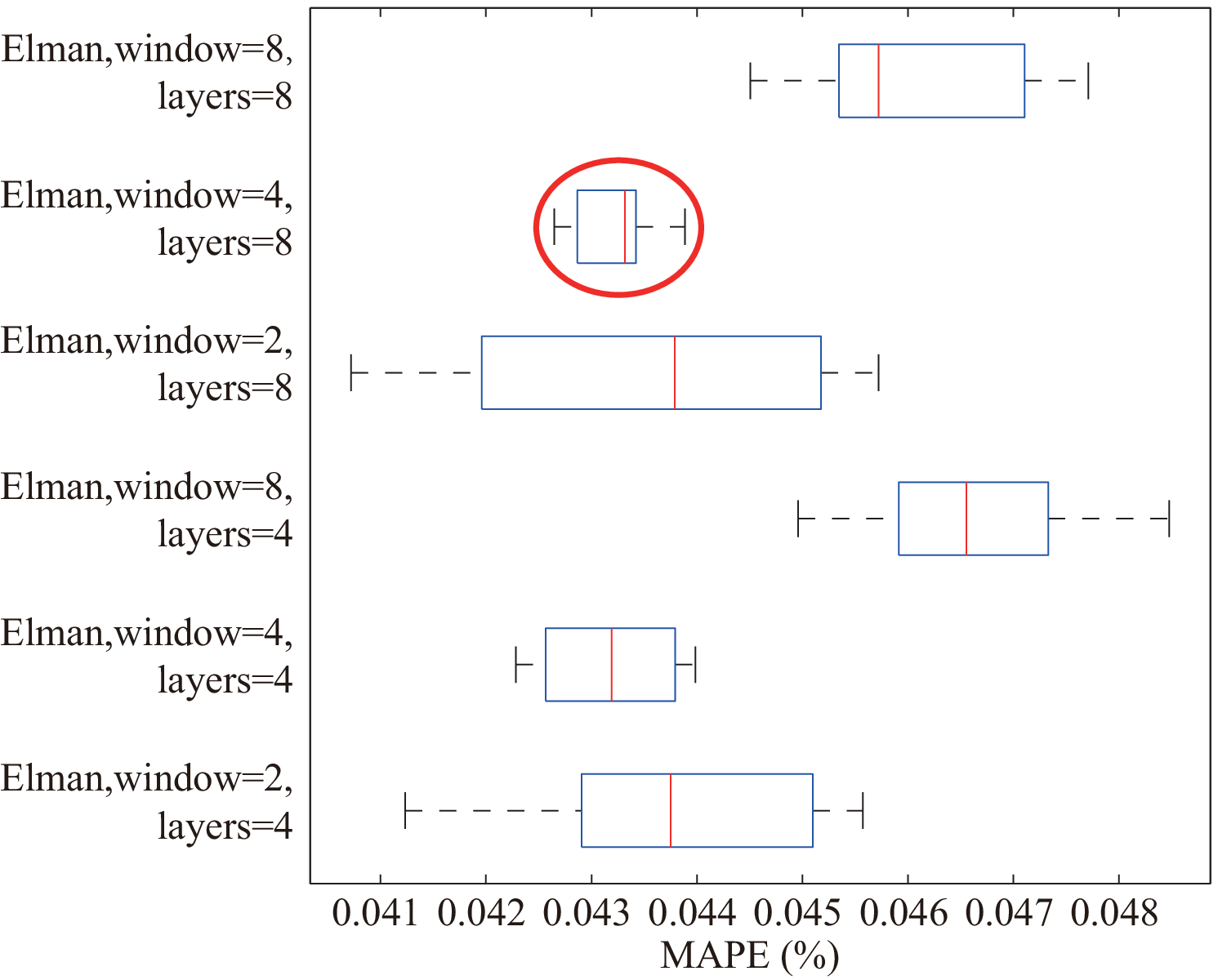}}
      \centerline{\tabincell{c}{(d) Models trained \\by using Dataset D}}\medskip
    \end{minipage}
    \vspace{-5mm}
    \caption{Boxplots of MAPE of each model trained by using 4 datasets.}
    \label{fig:boxplotdataamount}
\end{figure}

\begin{figure}[htb]
    \begin{minipage}[b]{0.48\linewidth}
      \centering
      \centerline{\includegraphics[width=0.95\linewidth]{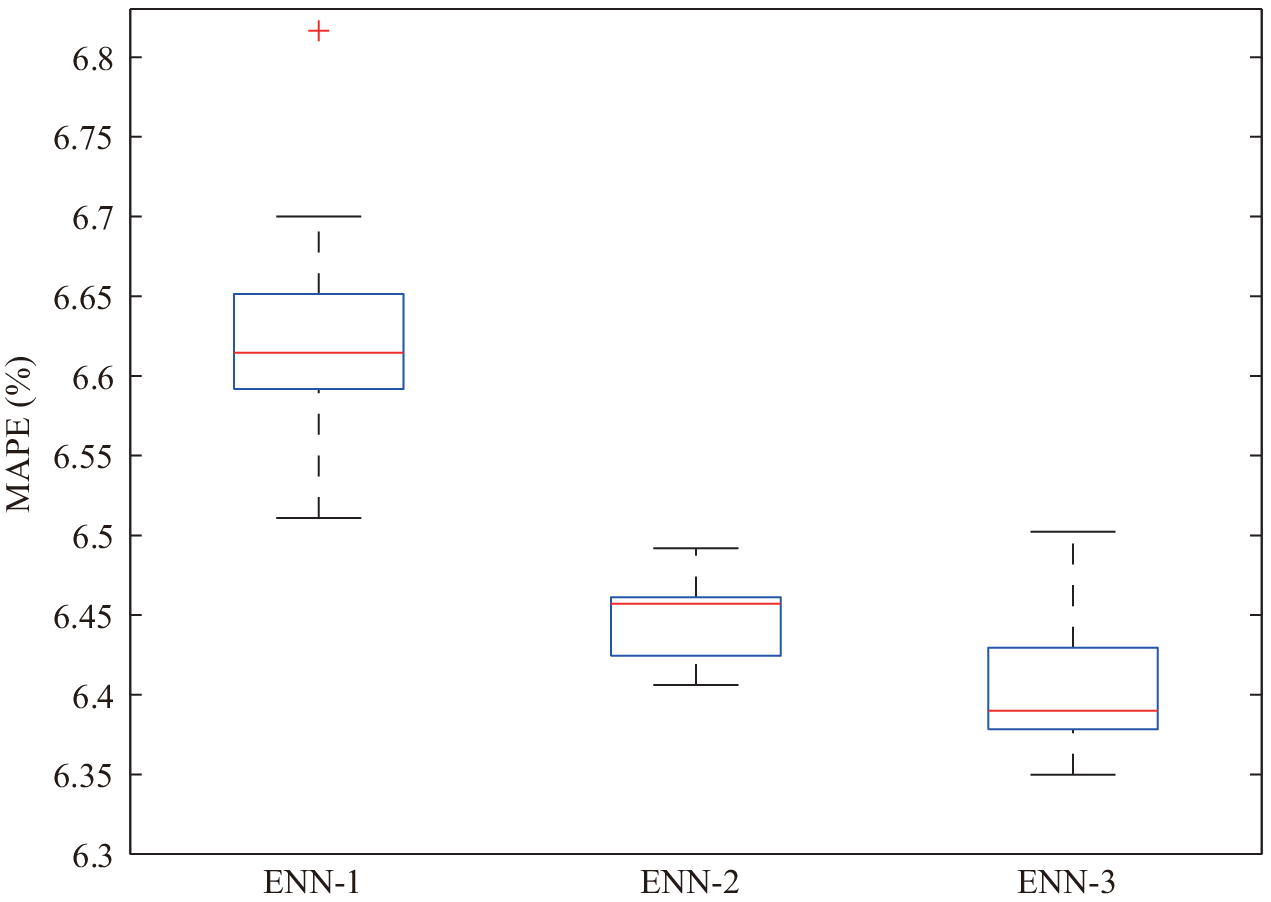}}
      \centerline{\tabincell{c}{(a) Models trained \\by using Dataset A}}\medskip
    \end{minipage}
    \hfill
    \begin{minipage}[b]{0.48\linewidth}
      \centering
      \centerline{\includegraphics[width=0.95\linewidth]{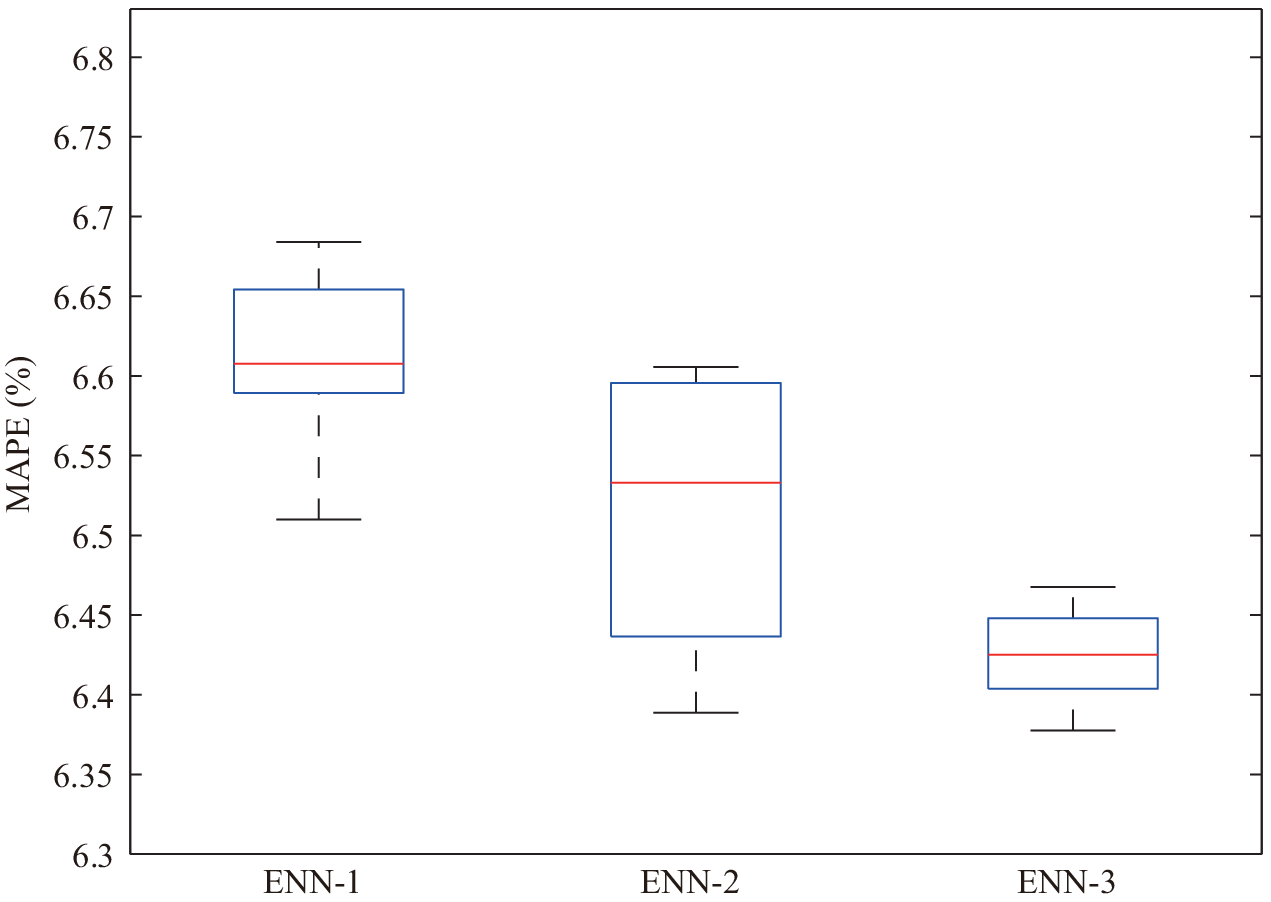}}
      \centerline{\tabincell{c}{(b) Models trained \\by using Dataset B}}\medskip
    \end{minipage}
    \begin{minipage}[b]{0.48\linewidth}
      \centering
      \centerline{\includegraphics[width=0.95\linewidth]{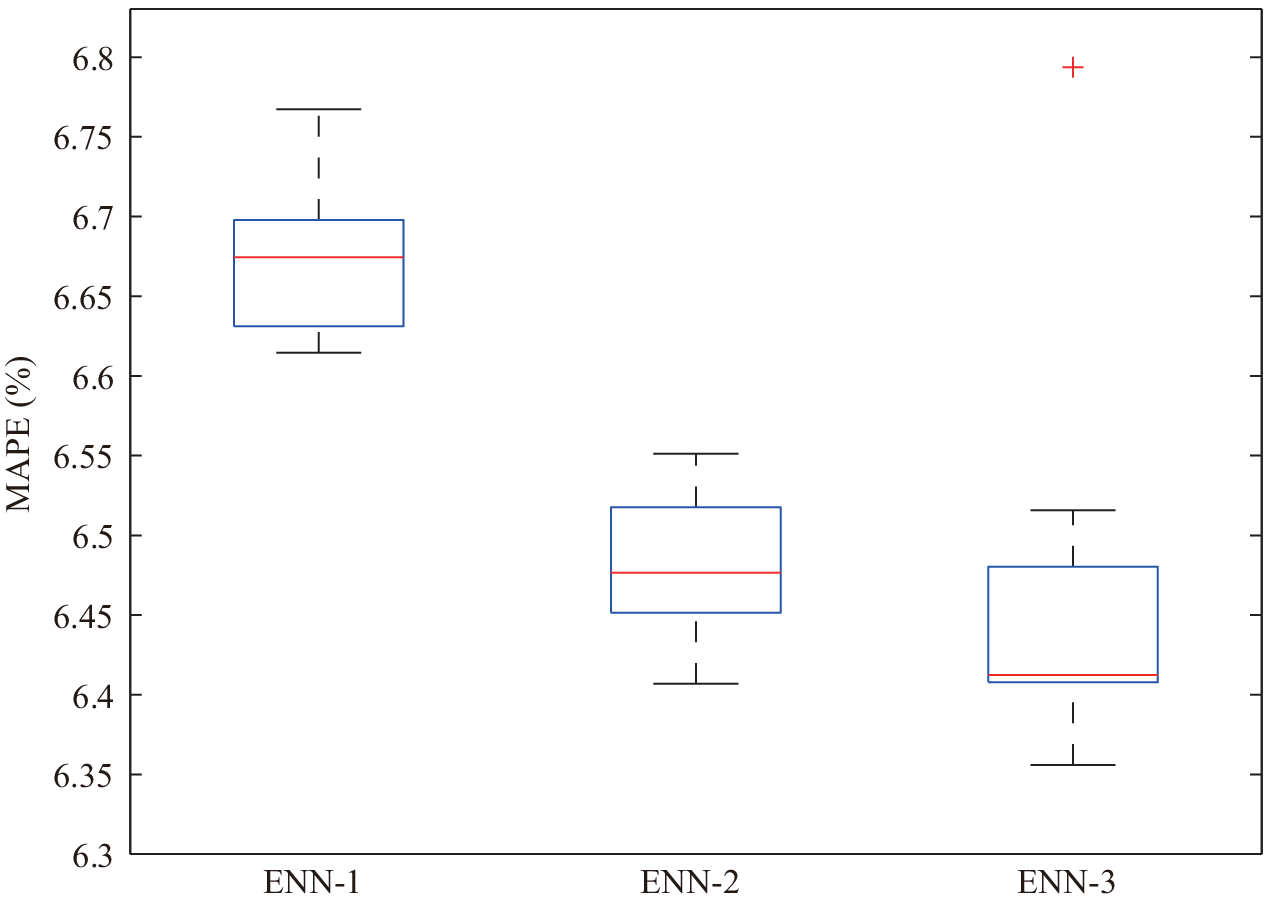}}
      \centerline{\tabincell{c}{(c) Models trained \\by using Dataset C}}\medskip
    \end{minipage}
    \hfill
    \begin{minipage}[b]{0.48\linewidth}
      \centering
      \centerline{\includegraphics[width=0.95\linewidth]{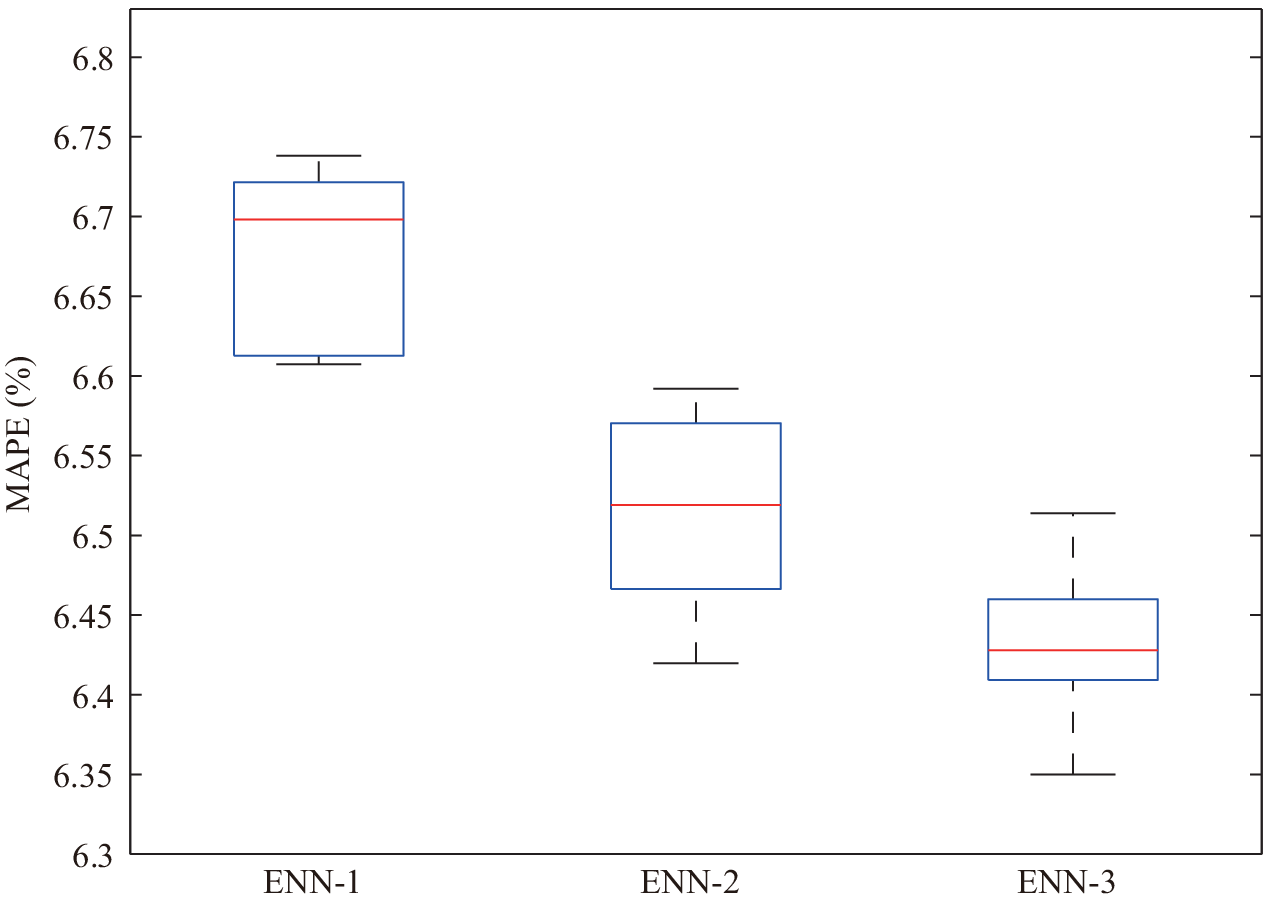}}
      \centerline{\tabincell{c}{(d) Models trained \\by using Dataset D}}\medskip
    \end{minipage}
    \vspace{-5mm}
    \caption{DMAPE of each model trained by using 4 datasets in a part of prediction interval.}
    \label{fig:dmapeofpart}
\end{figure}

Figure \ref{fig:boxplotdataamount} and \ref{fig:dmapeofpart} show the boxplots of MAPE and the tendency of daily MAPE (DMAPE) in a part of prediction interval trained by using 4 datasets. DMAPE shows daily tendency of prediction errors clearly
\begin{equation}
    \text{DMAPE}=\frac{1}{24}\sum_{i=1}^{24}\frac{|y_i-y_{pi}|}{y_i}\times100\%,\label{eq:dmape}
\end{equation}
\noindent where $y_i$ is the actual heat demand and $y_{pi}$ is the predicted result.

It is obvious that when more data have been used for model training, a better accuracy can be obtained.

Meanwhile, the results of the Student's \emph{t}-test (Table \ref{tab:diffdataamountp}) show that the distributions of the MAPE for the models trained with different amounts of data have significantly statistical difference. It means that increasing the amount of training set can improve the model performance and reduce MAPE and RMSE.

\begin{table}[htbp]
    \centering
    \caption{\emph{P}-value of the Student's \emph{T}-test between MAPE of Models}
    \label{tab:diffdataamountp}
    \vspace{-4mm}
    \begin{tabular}{ccccc}
        \hline
        \multirow{2}{2cm}{Models}&\multicolumn{4}{c}{Dataset used}\\
        &Dataset A&Dataset B&Dataset C&Dataset D\\
        \hline
        ENN-1 and ENN-2&2.17E-06&0.004428&4.29E-08&4.03E-06\\
        ENN-1 and ENN-3&2.65E-07&9.22E-09&8.78E-05&1.65E-09\\
        ENN-2 and ENN-3&0.009206&0.004532&0.584193&0.002252\\
        \hline
    \end{tabular}
\end{table}

According to the results of MAPE, RMSE, the Student's \emph{t}-test and boxplots, three year data were used for model training.

\subsection{The Impacts of Key Parameters}\label{ssec:impactofkeyfactors}

The model, ``Elman, window=4, layers=8'', trained by using Dataset A, Dataset B, Dataset C or Dataset D, was renamed as ENN-A, ENN-B, ENN-C and ENN-D. Correspondingly, they use the ambient temperature, ambient temperature + direct solar radiance, ambient temperature + wind speed, and ambient temperature + direct solar radiance + wind speed as inputs, respectively. Table \ref{tab:mapermsemad} lists MAPE and RMSE. Generally speaking, all models are capable to reflect the change of heat demand and predict the heat demand with MAPE less than 6.60\% and RMSE less than 14.8000. ENN-D shows the best accuracy with MAPE=6.35\% and RMSE=14.5358. It is also clear that the introduction of direct solar radiance and wind speed has positive impacts on the performance as ENN-B, C and D have smaller MAPE and RMSE than ENN-A. Comparatively, the inclusion of wind speed results in a better prediction accuracy than that of the direct solar radiance. This implies that wind speed is a more important parameter. Meanwhile, the introduction of both wind speed and direct solar radiance simultaneously can further improve the model accuracy.


Table \ref{tab:mapermsemad} also presents the maximum absolute deviation (MAD) of different models
\begin{equation}
    MAD=\max_i|y_i-y_{pi}|,\label{eq:mad}
\end{equation}
\noindent where $y_i$ is the actual heat demand and $y_{pi}$ is the predicted result.

It is clear that compared to ENN-A, including direct solar irradiance and wind speed can reduce MAD. Meanwhile, including direct solar irradiance is more effective than including wind speed to reduce MAD, even though including wind speed (ENN-C) can result in a lower MAPE.

\begin{table}[htbp]
    \caption{MAPE, RMSE and MAD of Different Models}
    \label{tab:mapermsemad}
    \vspace{-4mm}
    \centering
    \begin{tabular}{cccc}
        \toprule
        Model&MAPE&RMSE&MAD\\
        \midrule
        ENN-A&6.50\%&14.7751&91.6261\\
        ENN-B&6.47\%&14.6969&71.8131\\
        ENN-C&6.43\%&14.6139&81.5368\\
        ENN-D&6.35\%&14.5358&70.6640\\
        \bottomrule
    \end{tabular}
\end{table}

\begin{figure}
   \centering
        \includegraphics[width=0.95\linewidth]{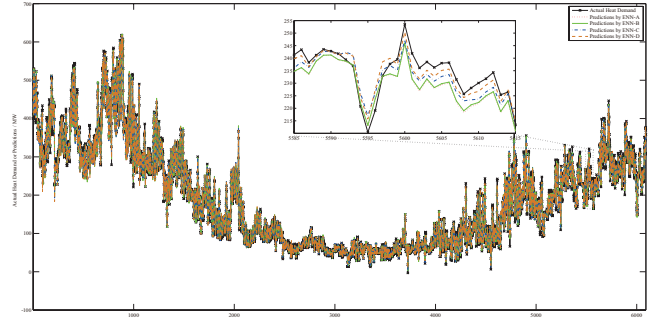}
   \vspace{-5mm}
   \caption{Distribution of absolute percentage errors.}\label{fig:distribofape}
\end{figure}

Figure \ref{fig:distribofape} shows the distribution of absolute errors (in percentage). For all models, most of errors ($>$74\%) is between -5\%$\sim$5\%. However, it is worth to note that although ENN-A has the highest MAPE, it has the most points in the range of -5\%$\sim$5\%. Meanwhile, there are more points which heat demand was under-estimated than those which heat demand was over-estimated for all of the models.

In order to further understand the error distributions, MAPE, RMSE and MAD of different models were re-calculated corresponding to different heat demand ranges and results were listed in Table \ref{tab:mapediffranges}-\ref{tab:maddiffranges}. For all of models, MAPE decreases, while RMSE increases with the increase of heat demands. However, direct solar irradiance and wind speed may have different influences at different heat demands. According to MAPE (Table \ref{tab:mapediffranges}), at very low demand (0$\sim$150MW), it is more beneficial to include wind speed (ENN-C); while in the demand of 150$\sim$300MW, including direct solar irradiance (ENN-B) has the lowest MAPE. In addition, despite that ENN-D has the lowest overall MAPE as shown in Table \ref{tab:mapermsemad}, it does not always have the lowest MAPE at different heat demands. It only has the lowest MAPE in the demand range of 300$\sim$450MW. In general, it can be concluded that the statistic model can clearly benefit from introducing more meteorological parameters at high demands, such as $>$450MW. It is mainly owing to that space heating accounts for a bigger fraction at higher demands and introducing more meteorological parameters can result in a better estimation of heat loss.

The influence of input factors on MAD is not consistent at different heat demands. As illustrated in Table \ref{tab:maddiffranges}, including direct solar irradiance can reduce MAD except in the demand range of 150$\sim$300MW and it is more effectively than including wind speed. Moreover, similar to the results about MAPE, including both direct solar irradiance and wind speed does not necessarily result in the lowest MAD.

\begin{table}[htbp]
    \caption{MAPE at Different Heat Demand Ranges}
    \label{tab:mapediffranges}
    \vspace{-4mm}
    \centering
    \begin{tabular}{ccccc}
        \toprule
        Heat demand (MW)&ENN-A&ENN-B&ENN-C&ENN-D\\
        \midrule
        0$\sim$150&8.95\%&8.87\%&8.73\%&8.83\%\\
        150$\sim$300&4.95\%&4.93\%&5.06\%&4.97\%\\
        300$\sim$450&4.14\%&4.13\%&4.06\%&4.04\%\\
        $>$450&3.89\%&3.71\%&3.63\%&3.66\%\\
        \bottomrule
    \end{tabular}
\end{table}

\begin{table}[htbp]
    \caption{RMSE at Different Heat Demand Ranges}
    \label{tab:rmsediffranges}
    \vspace{-4mm}
    \centering
    \begin{tabular}{ccccc}
        \toprule
        Heat demand (MW)&ENN-A&ENN-B&ENN-C&ENN-D\\
        \midrule
        0$\sim$150&9.8809&9.8798&9.9209&9.8010\\
        150$\sim$300&15.9006&15.9266&16.0522&15.9812\\
        300$\sim$450&19.8979&19.7661&19.2720&19.2544\\
        $>$450&25.2362&24.3013&23.7500&23.7959\\
        \bottomrule
    \end{tabular}
\end{table}

\begin{table}[htbp]
    \caption{MAD at Different Heat Demand Ranges}
    \label{tab:maddiffranges}
    \vspace{-4mm}
    \centering
    \begin{tabular}{ccccc}
        \toprule
        Heat demand (MW)&ENN-A&ENN-B&ENN-C&ENN-D\\
        \midrule
        0$\sim$150&28.5977&26.4968&26.5604&23.7552\\
        150$\sim$300&46.4071&55.1570&52.5413&54.5364\\
        300$\sim$450&91.6261&71.8131&81.5368&70.6640\\
        $>$450&48.5284&43.9760&46.3392&47.8068\\
        \bottomrule
    \end{tabular}
\end{table}

The distribution of absolute errors (in percentage) was also broken down at different heat demands and shown. Obviously, no matter what the heat demand is, the most of errors of all models are in the range of -5\%$\sim$5\%, which is similar to the overall error distribution. Comparatively, for the heat demand of 0$\sim$150MW, the fraction of the points with errors larger than 10\% or lower than -10\% is much higher. That is the reason that all models have the worst MAPE.
%

\subsection{Discussions}\label{ssec:discussions}

As shown in Table \ref{tab:mapediffranges}-\ref{tab:maddiffranges}, ENN may not benefit from the introduction of both wind speed and direct solar irradiance simultaneously. The potential reasons include: 1) the effect of one factor is already included into another (e.g., the effect of wind on the demand is reflected by the temperature); hence, it is not necessary to include redundant information in the analysis. 2) For ENN, there are different ways when integrating multi-factors. In order to further improve the ENN, advanced fusion methods may be applied. For example, with the principle of hierarchical neural network, two ENNs, each one with a single factor, can be trained separately. Then the two trained ENNs can be combined to get an ENN model for prediction. 3) Data discontinuity and outliers, including additive outlier, level shift, temporary change and ramp effect of data quality, may also have influence on the performance of heat demand prediction \cite{box75}. Therefore, there is a strong requirement about checking the data quality.

\section{Conclusion}\label{sec:conclusions}

Ambient temperature, wind speed and direct solar irradiance have been recognized as key input parameters in the development of statistic models for predicting heat demand. In order to understand the quantitative influence of different meteorological parameters, this paper developed a new model based on Elman neural network for the prediction of the heat demand from the heat production side. Based on the results, that the following conclusions are drawn:

\begin{itemize}
    \item ENN with multiple hidden layers has a good ability to predict the heat demand accurately; and the slide window and the number of layers have been identified as two key parameters influencing the model performance. With a slide window of 4 hours and a number of layers of 8, the mean absolute percentage error (MAPE) is less than 7\%.
    \item Including wind speed can result in a lower overall MAPE (6.43\%) than including direct solar irradiance (6.47\%); therefore, wind speed should receive more attention when developing a statistic model.
    \item Including both wind speed and direct solar irradiance can further improve the overall performance, which MAPE is 6.35\%.
\end{itemize}

\end{document}